\documentclass[sigconf]{acmart}
\usepackage[utf8]{inputenc}
\usepackage{afterpage}
\usepackage{listings}
\setcopyright{none}
\newcommand{\degree}{$^\circ$}
\begin{document}
%
\settopmatter{printacmref=false}
\renewcommand\footnotetextcopyrightpermission[1]{}
\pagestyle{plain}
\title{APES: a Python toolbox for simulating reinforcement learning environments}

\author{Aqeel Labash}
\orcid{0000-0002-2201-2809}
\affiliation{%
  \institution{University of Tartu}
}
\email{aqeel.labash@gmail.com}

\author{Ardi Tampuu}
\affiliation{%
  \institution{University of Tartu}
}
\email{ardi.tampuu@ut.ee}

\author{Tambet Matiisen}
\affiliation{%
  \institution{University of Tartu}
}
\email{tambet.matiisen@ut.ee}

\author{Jaan Aru}
\affiliation{%
  \institution{University of Tartu}
}
\email{jaan.aru@ut.ee}

\author{Raul Vicente}
\affiliation{%
  \institution{University of Tartu}
}
\email{raulvicente@gmail.com}

\begin{abstract}
Assisted by neural networks, reinforcement learning agents have been able to solve increasingly complex tasks over the last years. The simulation environment in which the agents interact is an essential component in any reinforcement learning problem. The environment simulates the dynamics of the agents' world and hence provides feedback to their actions in terms of state observations and external rewards. To ease the design and simulation of such environments this work introduces \texttt{APES}, a highly customizable and open source package in Python to create 2D grid-world environments for reinforcement learning problems. \texttt{APES} equips agents with algorithms to simulate any field of vision, it allows the creation and positioning of items and rewards according to user-defined rules, and supports the interaction of multiple agents. 

\end{abstract}
%
%


\maketitle

\section{Introduction}
Reinforcement learning is a type of machine learning in which an agent learns from its experience in an environment to maximize some cumulative reward \cite{sutton1998reinforcement,mnih2015human}. The environment is a crucial part of any reinforcement learning setting, as the environment defines the dynamics of the world in which the agent is embedded and hence it determines (possibly stochastically) the feedback to the agent's actions. This feedback includes the next observation state for the agent and its external rewards. 

Large part of creating a reinforcement learning task consists of programming the environment or world in which the agent will be immersed and interacting with other elements. This includes the properties of all possible items in the environment, their positioning, the set of events triggering rewards and punishments, and the computation of the next observation state. The latter can be complicated by the perceptual properties with which we would like to endow our agent. In particular, more realistic agents typically do not observe directly the environment state but just a function or partial state of it. For example, simulating the finite field of vision of an agent in an environment with occluders is a common situation leading to such partial observations of states.

In this work we present \texttt{APES} (Artificial Primate Environment Simulator), a Python tool to create 2D grid-world environments for reinforcement learning tasks. The name derives from the use of our environment for simulating real experiments with real apes competing for a reward in a controlled environment \cite{hare2000chimpanzees}. \texttt{APES} allows the user to quickly build 2D environments for reinforcement learning problems by allowing the user to add layers of different types of pre-defined or new items, set any rewarding scheme, and interface an agent with the environment (i.e. compute the agent's next observation state and perform the agent's action). Moreover, the agent can be equipped with different types of vision with customizable parameters. Importantly, \texttt{APES} also supports the interaction of multiple agents in the same environment, and hence could be used to train and test multiagent reinforcement learning algorithms \cite{busoniu2008comprehensive,schwartz2014multi,tampuu2017multiagent,foerster2016learning}. The full code can be found at \url{https://github.com/aqeel13932/APES}.

In the next section we describe some alternative software solutions. The rest of the paper is the core of this contribution and it explains the organization and capabilities of \texttt{APES} as well as an example of its use to simulate a simple reinforcement task. 


\subsection{Related Work}
There are other solutions to simulate simple environments for reinforcement learning tasks. Here, we focus on \texttt{MazeBase}, \texttt{OpenAI gym}, and \texttt{VGDL} and note their main differences with \texttt{APES}. \texttt{MazeBase} \cite{DBLP:journals/corr/SukhbaatarSSCF15} is a 2D grid-world environment to create simple games programmable by the user. \texttt{MazeBase} can also have a non-visual world representation. Although \texttt{MazeBase} offers a variety of types and co-existing elements on same grid cell, the agents have limited vision options. The wrapper \texttt{OpenAI gym} \cite{DBLP:journals/corr/BrockmanCPSSTZ16} offers numerous games but lacks the option to customize the environment, and most games are fully observed environments.
\texttt{VGDL}\cite{schaul2013pyvgdl} allows to define different kinds of games in declarative language, yet it lacks the control of some vision properties such as partial observability in 2D.

\section{Environment}
\begin{figure}[t]
\begin{center}
\includegraphics[scale=0.5]{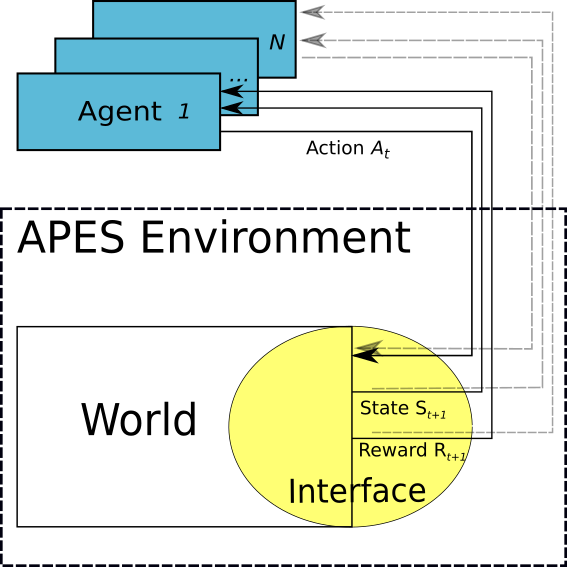}
\caption{\texttt{APES} environment and its relation to the framework of reinforcement learning. \texttt{APES} handles the world environment and its interface with a number of agents controlled by external algorithms.} \label{fig.env}
\end{center}
\end{figure}

\texttt{APES} is organized around two basic structures: the World and the agents' interface with the world. The first defines the characteristics and dynamics of the environment, while the second describes the interaction of one or multiple agents with the environment. Each agent can be controlled with any algorithm external to \texttt{APES} (say your favourite implementation of deep reinforcement learning) and the agent solely interacts with \texttt{APES} by receiving an observation state and reward, and sending an action to be performed in the world. See Figure \ref{fig.env} for a diagram illustrating the APES environment within the framework of reinforcement learning. Next we describe the elements of these structures. We begin with the world.

\subsection{World}
The world arena is a 2D grid where all elements are located and interact. In Figure \ref{example} we illustrate a snapshot of a simple environment created with \texttt{APES}. To define the properties of the environment and the elements therein the user can specify each of the following features:
\begin{itemize}
\item \textbf{World size:} dimensions of the 2D grid defining the arena. Current version supports only square-shaped environments.
\item \textbf{End of episode:} the world has the attribute \texttt{Terminated} to identify when an episode ends. The user can specify an episode to end in any of these cases (or whichever occurs first): 
\begin{itemize}
\item the maximum number of time steps has been reached (specified by \textbf{Maximum steps}).
\item all rewards have been consumed.
\end{itemize}
After the flag \texttt{Terminated} has being raised the environment will be regenerated.

\begin{figure}[t]
\centering
\includegraphics[scale=0.4,trim={13mm 8mm 0mm 0mm},clip]{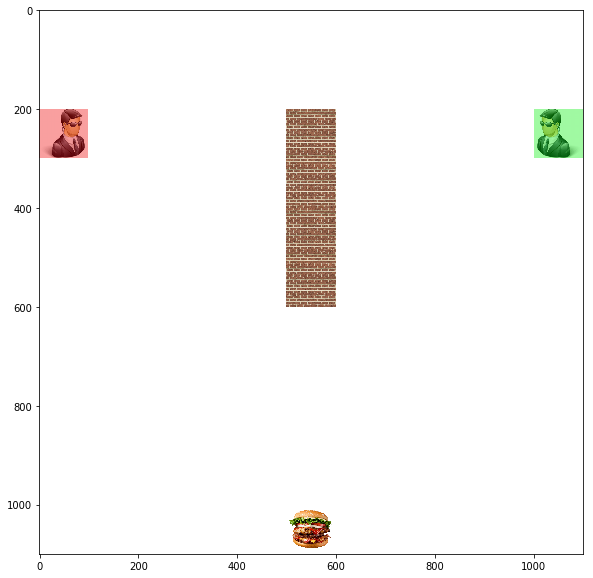}
\caption{Example of a simple grid world environment populated by two agents, an elongated vertical wall and a food item.}
\label{example}
\end{figure}



\item \textbf{Items:} elements to be located at the cells of the 2D grid. There are different types of pre-defined items:
\begin{itemize}
    \item \textbf{Food:} item reachable by agents. Agents can occupy the same cell as the food item and collect it, making it disappear from the environment. Although typically associated with a positive reward, it can be associated with any type or reward or punishment according to the Rewarding scheme defined by the user.
    \item \textbf{Obstacles:}
    items responsible for stopping agents from moving freely in the World. An obstacle element can be chosen from the following list. 

    \begin{itemize}
    \item \textbf{Wall:} obstacles that cannot be traversed and impede visibility through them.
    \item \textbf{Water:} obstacles that cannot be traversed but they allow visibility through them.
    \end{itemize}
    
    Each obstacle can be associated with a \textbf{Shape} matrix determining the geometric shape of the obstacle. For example, adding a "T"-shaped obstacle can be done by passing 3$\times$3 array with first row and middle column as ones and the rest as zeros.

    \item \textbf{Agent:} interactive elements in the environment which are controlled by external algorithms. They interact with the items in the World by executing actions and receiving observation states and rewards. Each agent can be associated a \textbf{Power} level to define the interaction with other agents. For example, in case of collision, agents of lower level can be set to receive some punishment. Each agent also has an attribute \textbf{Transparency} to make the agent act as a vision barrier or as a transparent element for other agents' vision.
    
    
\newcommand\x{0.21}
\begin{figure*}
\includegraphics[scale=\x]{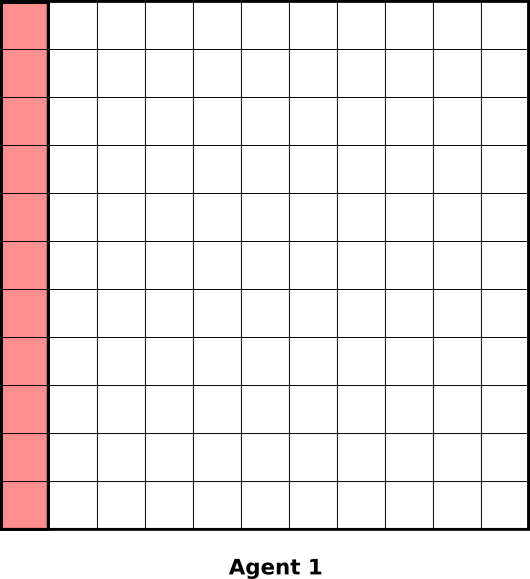}
\includegraphics[scale=\x]{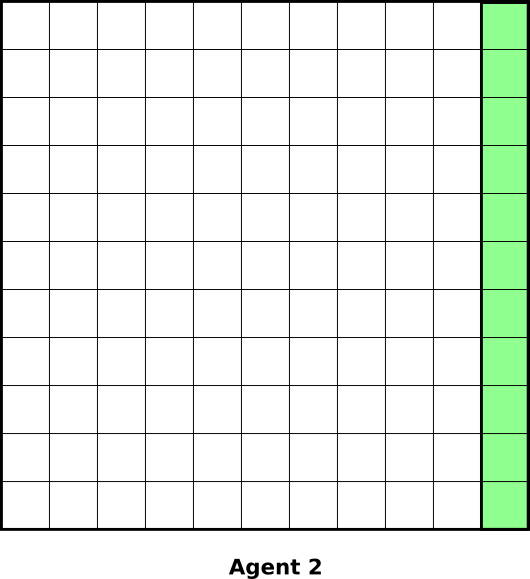}
\includegraphics[scale=\x]{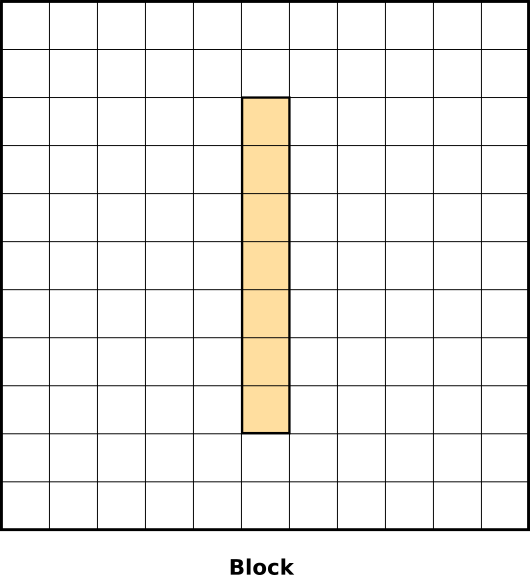}
\includegraphics[scale=\x]{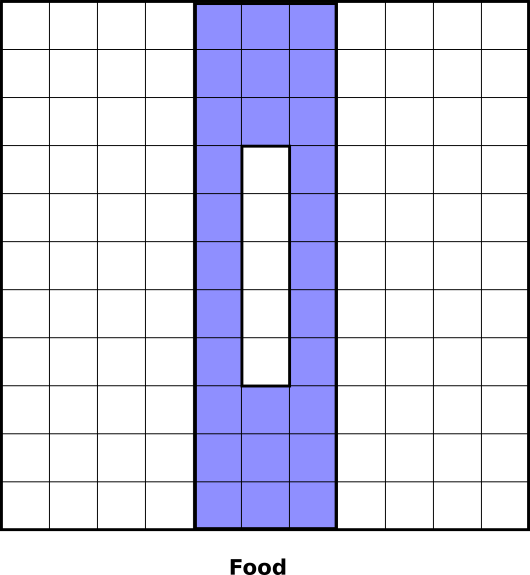}
\includegraphics[scale=\x]{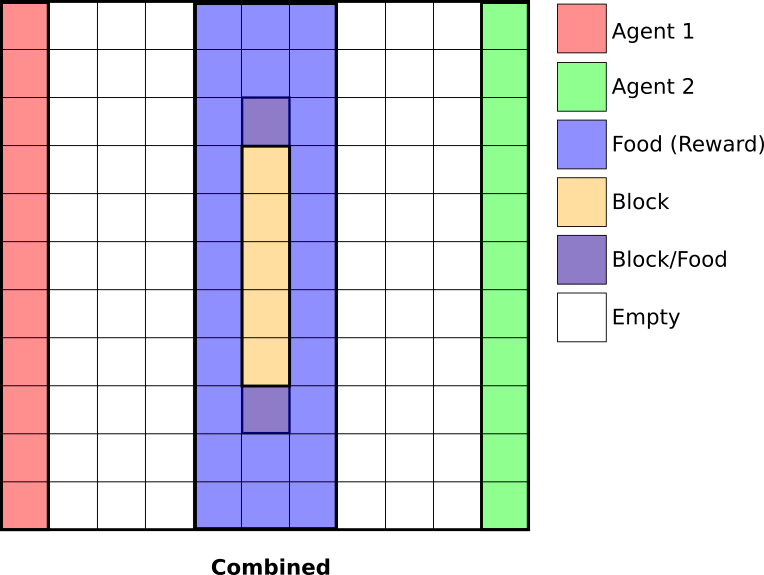}\\
\caption{Example of probability distribution matrices for different elements in the environment. Each episode is regenerated by sampling each element's location from their associated probability distribution.} \label{fig.prob}
\end{figure*}

\end{itemize}
    \item \textbf{Probability Distribution Matrix (PDM):} matrix of same size as the world grid used to distribute elements across it. The matrix contains the probability of an element to be located at each position at the start of each episode. Figure \ref{fig.prob} illustrates a graphical representation of PDMs used to distribute several elements randomly within some defined locations. PDM values will be automatically normalized and multiple elements can have the same PDM. If no PDM is given, elements are placed randomly across the entire grid by default. When two elements spawn in the same position the placing priority is: 1) agent, 2) food, and 3) obstacle. For obstacles with  more than one block, only the center is guaranteed to appear. Extensions from the center of the obstacle will be overruled by other elements if such conflict occurred. 


\item \textbf{Block size:} size of the images of each element. This feature is used only for visualization of the environment as pictures or video.

\end{itemize}

\subsection{Agents' Interface}
The agents' interface defines the information communicated between the environment and each agent. It consists of the observation state and reward received by the agent at every time step, as well as the execution of the action chosen. Next we describe the content of each of these streams per agent and how they are computed.

\medskip

\textbf{Observation:}  determines what the agent obtains as observation depending on the environment state and vision algorithm. At each time step the observation fed back to the agent contains:

\begin{itemize}
\item \textbf{Observability:} a binary 2D array of same size as the world with each cell marked as $1$ if the corresponding cell location is observable by the agent.
\item \textbf {Food:} a binary 2D array of the same size as the world with each cell marked as $1$ if a food item is observed at the corresponding cell location.
\item \textbf{Agent position:} a binary 2D array of same size as the world size with a $1$ in the cell currently occupied by the focal agent.
\item \textbf{Agent orientation:} one hot vector of size 4 (North, South, East, West) with a $1$ indicating the current orientation of the focal agent.
\item \textbf {Other agent's position:} a binary 2D array per agent (other than the focal agent) of same size as the world and with a $1$ in the cell currently occupied the agent. If the agent is not currently observed by the focal agent the array will be all zeros.  
\item \textbf{Other agent's orientations:} one hot vector of size 4 (North, South, East, West) per agent (other than the focal agent) with a $1$ indicating the current orientation of the agent. If the agent is not currently observed by the focal agent the array will be all zeros.
\end{itemize}

\medskip

\textbf{Agent Vision:} it determines how and which parts of the environment state are transformed to produce an observation state for the agent. APES simulates two modes of agent's vision.  

\begin{figure*}[ht!]
\centering
\includegraphics[scale=0.45,trim={10mm 6mm 280mm 4mm},clip]{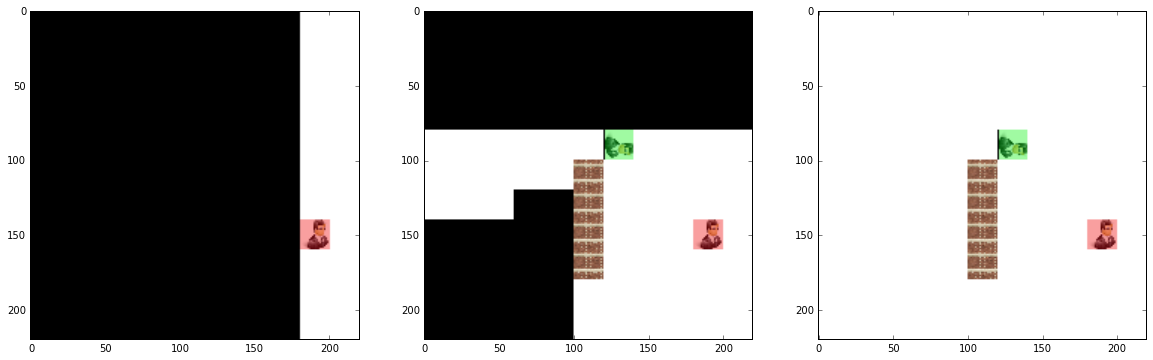}
\hspace{10px}
\includegraphics[scale=0.45,trim={150mm 6mm 140mm 4mm},clip]{Images/centric.png}
\caption{The left panel shows the field of vision of an agent (red) with allocentric vision with 180{\degree} vision angle and unlimited vision range. The right panel illustrates the totality of the world environment.}
\label{centric}
\end{figure*}

\begin{figure*}
\centering
\includegraphics[scale=0.45,trim={150mm 6mm 140mm 4mm},clip]{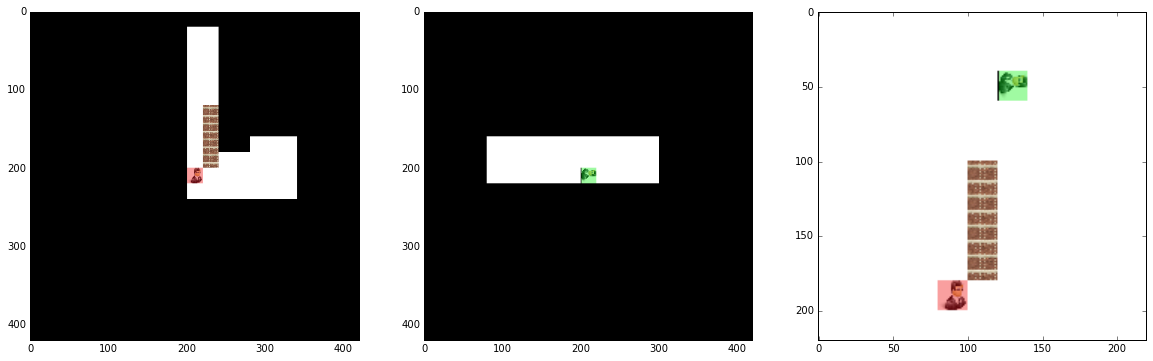}
\hspace{10px}
\includegraphics[scale=0.45,trim={288mm 6mm 0mm 4mm},clip]{Images/egocentric.png}
\caption{The left panel shows the field of vision of an agent (green) with egocentric vision with 180{\degree} vision angle and unlimited vision range. The right panel illustrates the totality of the world environment.}
\label{egocentric}
\end{figure*}

\begin{itemize}

\item \textbf{Allocentric:}\label{agnt:cent}
the agent receives a bird's eye view of the environment. As the agent executes actions and moves across the environment the agent's position in the visual field will change depending on its location as shown in Figure \ref{centric} where the black area represents the unobserved area.

\item \textbf{Egocentric:}\label{agnt:alter}
provides a more natural type of vision in which the field of view is centered at agent's location. Hence the agent will always be in the center of the view and the surrounding will change depending on its location and direction. Figure \ref{egocentric} illustrates the environment state from this point of view.

\end{itemize}

\begin{figure*}
\centering
\includegraphics[scale=0.45,trim={10mm 6mm 280mm 4mm},clip]{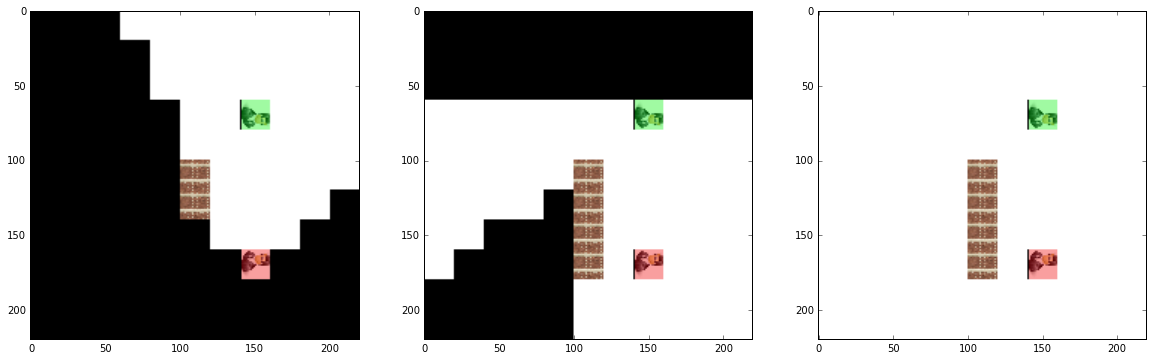}
\hspace{10px}
\includegraphics[scale=0.45,trim={150mm 6mm 140mm 4mm},clip]{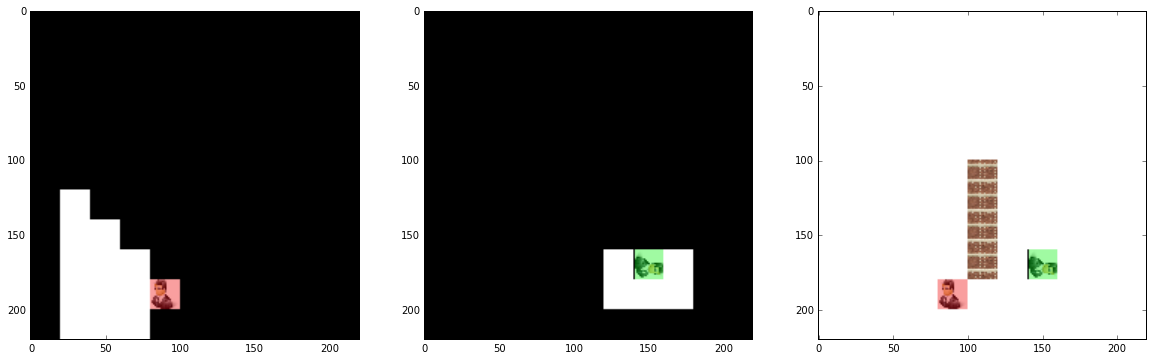}
\caption{Left panel shows an agent (red) with 90{\degree} vision angle with unlimited range. Right panel shows an agent (green) with a vision range of one unit and vision angle of 180{\degree}.}
\label{visionangle_limit}
\end{figure*}

For both modes APES uses a shadow-casting algorithm to determine the agent's field of vision based on the elements populating the 2D grid \cite{fov}. Each of the two modes of vision can be tuned by parameters such as:

\begin{itemize}

\item \textbf{Vision Angle:}\label{agnt:va}
the angle spanning the agent's field of view. The value should be between 0 to 360. Left panel in Figure\ref{visionangle_limit} shows the effect of the vision angle on an agent's field of view.
\item \textbf{Vision Limit:}\label{agnt:vl} extent of the agent's field of view. It is the maximum distance the agent can observe. If set to value -1 the agent obtains an infinite vision limit. Right panel in Figure\ref{visionangle_limit} shows the effect of the vision range on an agent's field of view. 

\end{itemize}

\medskip

\textbf{Reward:} provides feedback to agent's actions by specifying the rewards and punishments obtained in case of different events occurring in the environment. The default option is a rewarding scheme given the vector [-10, 10, -0.1] indicating a value of -10 for the event of colliding with another agent, +10 for finding food item, and -0.1 for every time step. Any rewarding scheme can be specified for any set of events defined by the user by passing a new reward function to \texttt{World class} instance.

\medskip





\textbf{Actions:} defines how the agent interacts and influences the environment, including other agents. The actions are selected according to the agent's controller which is usually an algorithm external to APES (although see the section on Autopilots for some automated agents provided by \texttt{APES}). The space of actions supported by \texttt{APES} and their ordering is given as follows:  

\begin{itemize}
\item \textbf{Set of actions:} there are two sets of primitive actions for each agent, 4 actions for looking in different directions (North, South, East, West) and another 4 actions for moving. Lastly, there is a no-operation action to complete a set of 9 primitive actions. These primitives can be concatenated to create different single complex actions, for example "Look left, Move Left" as one action which can be executed in one time step.

\item \textbf{Actions Order:} actions of different agents are executed in temporal order according to an ordered array of the agents' identifiers. 


\end{itemize}


\begin{figure}
\centering
\includegraphics[scale = 0.3]{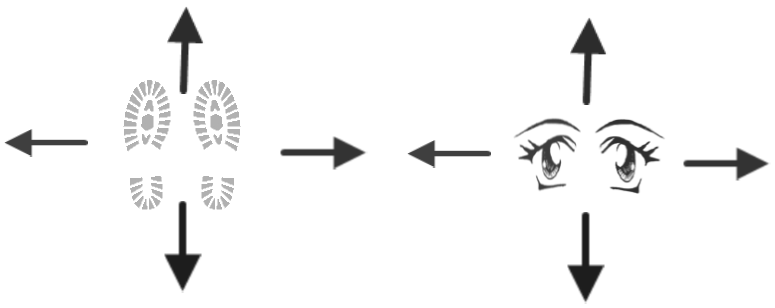}
\caption{The elementary actions available for the agent are: moving (north, south, east, west), look (north, south, east, west), and no operation.}\label{AgentAction}
\end{figure}



\subsection{Autopilots}
The design of the agents' controllers is a task separated from the \texttt{APES} environment. Any agent controller can interface with \texttt{APES} environment by reading the next observation state and reward and outputting an action within the space of actions described above. How the controllers make use of the environment state and reward to select the next action is upon the user's preferred reinforcement learning algorithm. However, to ease the interaction with some simple behaving agents \texttt{APES} includes the following autopilots:

\begin{itemize}
\item \textbf{Random Walker:} this autopilot randomly assigns actions to the agent regardless of its position or vision. For example, it can select the action "Move Left" although there is a wall on the left.

\item \textbf{Algorithmic Agent:} this autopilot has two properties: it has incremental vision (retains within its field of view all elements once seen) and navigates with a smart pathfinding algorithm. In particular, this agent randomly explores the unobserved area but once a food item is spotted the agent takes the shortest path to it according to the A* pathfinding algorithm \cite{hart1968formal}.
\end{itemize}


\subsection{Performance}
The performance in executing updates of the environment is a crucial issue since if not properly optimized this time can become a bottleneck in the training and testing of reinforcement learning algorithms. The speed depends on the number of the agents and the size of the world. For two agents and a world of size 11$\times$11 it can perform around 450 steps per second in a Intel Core i5-5200U CPU.

\section {Example}
\begin{figure}[b]
\centering
\includegraphics[scale=0.5,trim={0mm 0mm 0mm 0mm},clip]{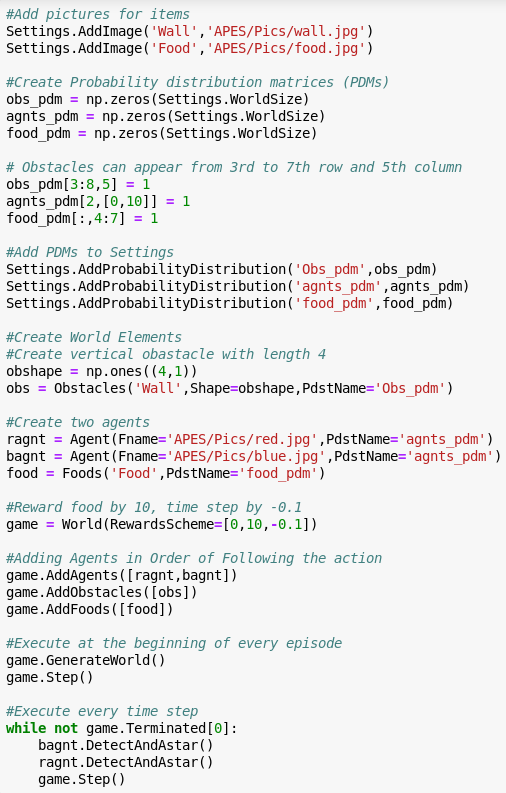}
\caption{The code to generate the environment illustrated in Figure \ref{example} and to set up a competition task between agent red and green.}
\label{code}
\end{figure}

Here we illustrate how \texttt{APES} can be set up to define and simulate a simple reinforcement learning task. Figure \ref{code} shows the code used to define and update the environment illustrated in Fig.\ref{example}. In this case at the beginning of each episode several elements (obstacles, agents, food) will be located randomly within some range of cells of the 2D-grid given by the probability distribution matrices. Given the rewarding scheme used in the example code the two agents (red and green) will compete for the food reward and will be punished every time step with a small penalty. 
Executing "world.Step()" after assigning each agent an action will update the environment for one time step. The regeneration of the environment (new episode) after the termination of the current episode is achieved by executing "Generate World".

\section{Conclusions}
In this article we presented the most important features in  \textit{APES} to design and simulate reinforcement learning tasks in 2D grid-worlds. In particular, \textit{APES}
handles the properties of the 2D arena and the elements therein. \textit{APES} also provides the feedback for the agents and executing their actions in the environment. Importantly, \textit{APES} supports different vision algorithms and multiple agents, which can make it attractive for studying multiagent reinforcement learning problems with partial observability. It is our hope that \textit{APES} can ease the creation and simulation of simple environments so that users can focus on the design and testing of better reinforcement learning algorithms.


\begin{acks}
A.L. was funded by the European Regional Development Fund and IT Academy. 
All authors thank the financial support from the Estonian Research Council through the personal research
grant PUT1476. This work was also supported by the Estonian Centre of Excellence in IT
(EXCITE), funded by the European Regional Development Fund. 
The authors are also indebted to Daniel Majoral  and Axel Harpe for taking time in working with the environment and reporting bugs.

\end{acks}

\bibliographystyle{ACM-Reference-Format}
\bibliography{acmart} 
\end{document}